\pdfoutput=1
\documentclass[11pt]{article}

\usepackage[final]{acl}
\usepackage{times}
\usepackage{latexsym}
\usepackage[utf8]{inputenc}

\usepackage{microtype}

\usepackage{inconsolata}
\usepackage{graphicx}
\usepackage{amssymb, algorithm, algpseudocode, algorithmicx, lipsum, xspace,multirow,booktabs,array,colortbl,multicol}
\definecolor{mydarkblue}{rgb}{0, 0, 0.5}
\usepackage{url}
\usepackage{tabularx}   % For tables with adjustable-width X columns
\usepackage{ragged2e}   % For \RaggedRight in X columns
\usepackage{listings}
\usepackage{times}
\usepackage{latexsym}
\usepackage{amsmath}
\usepackage[capitalize,noabbrev]{cleveref}
\usepackage{booktabs}
\usepackage{xcolor}
\usepackage{enumitem}
\usepackage{tcolorbox}
\usepackage{array}
\usepackage{booktabs}
\usepackage{csquotes}
\usepackage{graphicx}
\usepackage{amsmath,amsfonts,bm}

\newcommand{\godel}{G\"odel\xspace}
\newcommand{\framework}{Gödel Agent\xspace}

\lstdefinestyle{pythonstyle}{
    language=Python,
    basicstyle=\ttfamily\small,
    keywordstyle=\color{blue},
    commentstyle=\color{green!50!black},
    stringstyle=\color{red},
    showstringspaces=false,
    numbers=left,
    numberstyle=\tiny\color{gray},
    frame=single,
    breaklines=true,
    tabsize=4,
}

\def\eqref#1{equation~\ref{#1}}
\def\1{\bm{1}}

\DeclareMathAlphabet{\mathsfit}{\encodingdefault}{\sfdefault}{m}{sl}
\SetMathAlphabet{\mathsfit}{bold}{\encodingdefault}{\sfdefault}{bx}{n}

\title{\framework{}: A Self-Referential Agent Framework for Recursively Self-Improvement}

\author{ 
 Xunjian Yin$^{\spadesuit}$\hspace{0.5mm},
 Xinyi Wang$^{\clubsuit}$\hspace{0.5mm},
 Liangming Pan$^{\diamondsuit}$\hspace{0.5mm}, 
 Li Lin$^{\spadesuit}$\hspace{0.5mm}
 \\
 \bf{Xiaojun Wan$^\spadesuit$\hspace{0.5mm},
 William Yang Wang$^\clubsuit$}
\hspace{0.2mm}\hspace{1.5mm} \\
$^\spadesuit$ Peking University \quad$^\clubsuit$University of California, Santa Barbara  \quad $^\diamondsuit$ University of Arizona \\
 \texttt{\{xjyin,wanxiaojun\}@pku.edu.cn} \quad \texttt{william@cs.ucsb.edu} 
}

\begin{document}
\maketitle

\begin{abstract}
The rapid advancement of large language models (LLMs) has significantly enhanced the capabilities of agents across various tasks. However, existing agentic systems, whether based on fixed pipeline algorithms or pre-defined meta-learning frameworks, cannot search the whole agent design space due to the restriction of human-designed components, and thus might miss the more optimal agent design. In this paper, we introduce \framework, a self-evolving framework inspired by the \godel machine, enabling agents to recursively improve themselves without relying on predefined routines or fixed optimization algorithms. \framework leverages LLMs to dynamically modify its own logic and behavior, guided solely by high-level objectives through prompting. Experimental results on multiple domains demonstrate that implementation of \framework can achieve continuous self-improvement, surpassing manually crafted agents in performance, efficiency, and generalizability. 
% While \method showcases the feasibility, challenges remain in achieving stable and scalable recursive optimization, which requires the collective effort of the entire research community. 
\end{abstract}

\section{Introduction}
As large language models (LLMs) ~\citep{openai2024gpt4technicalreport,dubey2024llama3herdmodels} demonstrate increasingly strong reasoning and planning capabilities, LLM-driven agentic systems have achieved remarkable performance in a wide range of tasks~\citep{Wang_2024}. 
Substantial effort has been invested in manually designing sophisticated agentic systems using human priors in different application areas. Recently, there has been a significant interest in creating self-evolving agents, that not only greatly reduce human labor but also produce better solutions. Given that human effort can only cover a small search space of agent design, it is reasonable to expect that a self-evolving agent with the freedom to explore the full design space has the potential to produce a more optimal solution.

\begin{figure}[t]
\setlength{\abovecaptionskip}{-0.1cm}
\setlength{\belowcaptionskip}{-0.4cm}
\begin{center}
%\framebox[4.0in]{$\;$}
% \fbox{\rule[-.5cm]{0cm}{4cm} \rule[-.5cm]{4cm}{0cm}}
\includegraphics[width=0.4\textwidth]{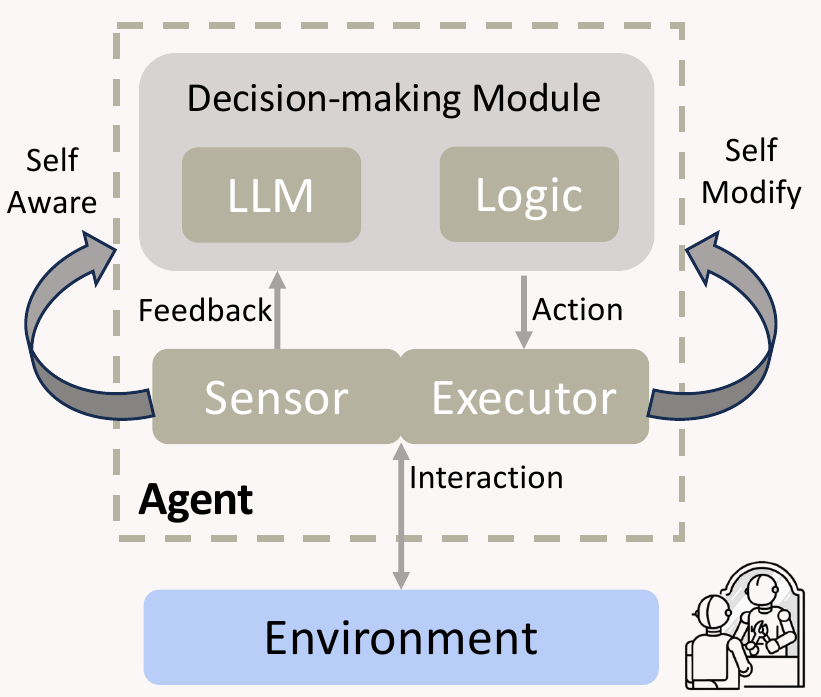}
\end{center}
\caption{Modular demonstration of \framework{}. Compared with traditional agents, its sensor and executor can read and write all of its own code. 
}
\label{fig:module}
\end{figure}

\begin{figure*}[t]
\setlength{\abovecaptionskip}{-0.1cm}
\setlength{\belowcaptionskip}{-0.4cm}
\begin{center}
%\framebox[4.0in]{$\;$}
% \fbox{\rule[-.5cm]{0cm}{4cm} \rule[-.5cm]{4cm}{0cm}}
\includegraphics[width=0.9\textwidth]{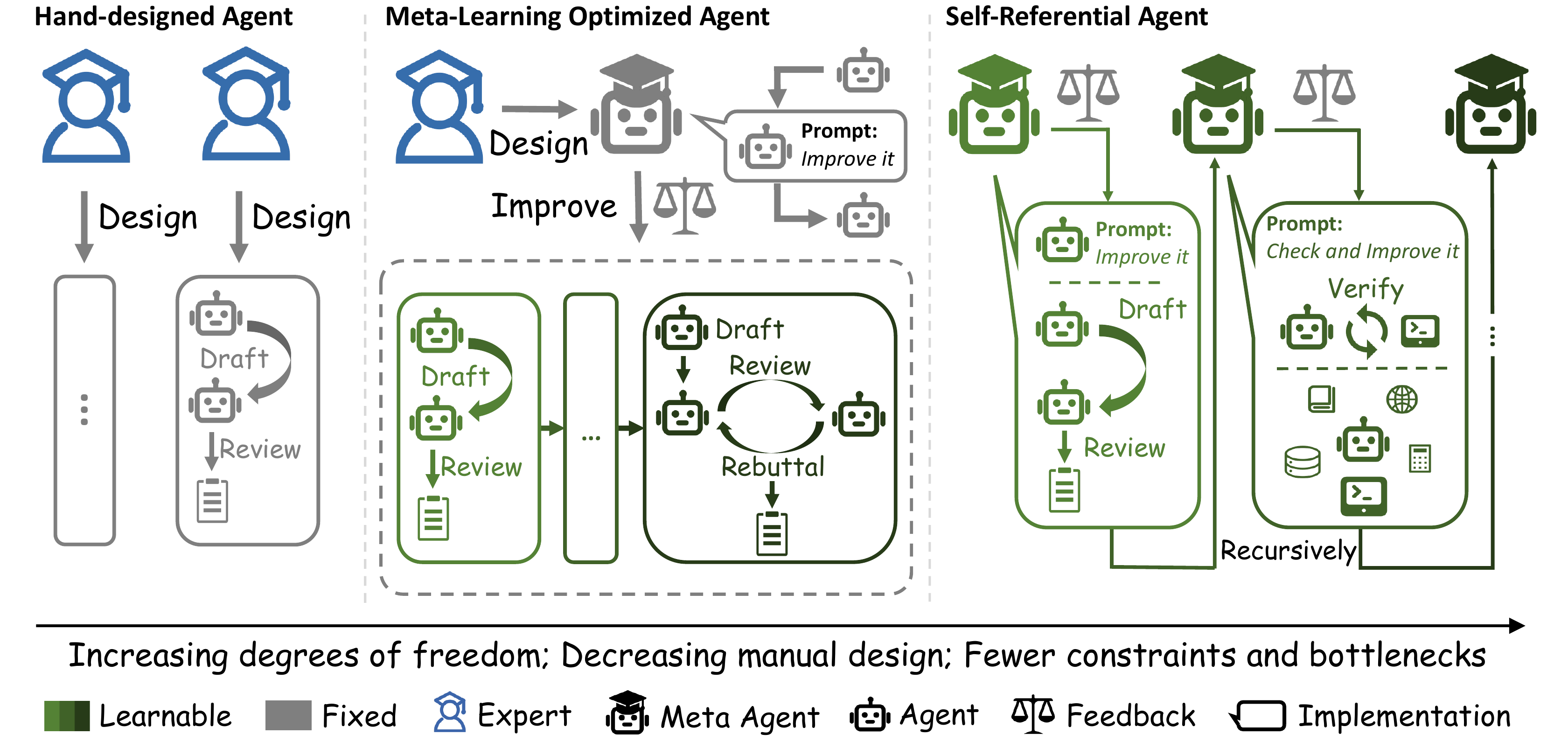}
\end{center}
\caption{Comparison of three agent paradigms. Hand-designed agents rely on human expertise which are limited in scope and labor-intensive. Meta-learning optimized agents are constrained by a fixed meta-learning algorithm, restricting their search space and optimization potential. In contrast, self-referential agent (\framework) can \textbf{recursively} improve itself without any limitation. 
Its optimization capabilities are constantly being enhanced by itself. Consequently, in return, it can continue to optimize itself better.
% Note that the input to \framework is itself, allowing it to modify itself and output a new version of itself.
}
%\wxy{This is too hard to understand as a first figure. Please see comments in the figure compare\_xinyi.pdf. I agree with Liangming that a more concrete example is needed. Perhaps you can use a simple task as a running example, and solve it with 3 different level agents?}}
\label{fig:compare}
\end{figure*}

There is a large body of work proposing agents capable of self-refinement. Some agents are designed to iterate over a fixed routine consisting of a list of fixed modules, while some of the modules are capable of taking self- or environment feedback to refine their actions ~\citep{chen2023teachinglargelanguagemodels, qu2024tool, tang2025chemagentselfupdatinglibrarylarge}. This type of agent, referred to as \textbf{Hand-Designed Agent}, is depicted as having the lowest degree of freedom in \cref{fig:compare}. More automated agents have been designed to be able to update their routines or modules in some pre-defined meta-learning routine, for example, natural language gradients~\citep{zhou2024symbolic}, meta agent~\citep{hu2024automated}, or creating and collecting demonstrations ~\citep{khattab2023dspy}. This type of agent, known as \textbf{Meta-Learning Optimized Agents}, is depicted as having the middle degree of freedom in \cref{fig:compare}.
However, there are inevitably some human priors involved in these agent designs that cannot be improved during the inference time.

% such as reflection~\citep{shinn2024reflexion}, self-debugging~\citep{chen2023teachinglargelanguagemodels}, tool learning~\citep{qu2024tool}, and search algorithms~\citep{yao2023treethoughtsdeliberateproblem}. 
% \plm{[expand this part] Once these agents are designed, they typically operate within fixed frameworks, which constrains their ability to adapt and improve.} \yxj{mention figure 1}

% In response, some works have proposed using learning algorithms to optimize agents, such as accumulating experiences in external memory~\citep{zhang2024survey}, optimizing prompts based on environmental feedback~\citep{khattab2023dspy}, refining agent logic through natural language gradients~\citep{zhou2024symbolic}, or using LLMs to design new agents automatically~\citep{hu2024automated}. 
% While these learning algorithms can optimize agents, they are also based on human priors and are fixed, which can be a bottleneck in the improvement of agents. \plm{what are the bottlenecks? give more details / examples.} \yxj{mention figure 1}

% It is evident that both types of agents above are inherently constrained by human priors and one intuitional method to further increase the freedom of self-improvement is to design a meta-meta-learning algorithm, to learn the meta-learning algorithm. However, there is always a higher-level meta-learning algorithm that can be manually designed to learn the current-level meta-learning method, creating a never-ending hierarchy of meta-learning. 

% \plm{Figure 1: a more concrete example is needed.} 

In this paper, we propose \textbf{\framework} to eliminate the human design prior, which is an automated LLM agent that can freely decide its own routine, modules, and even the way to update them. 
It is inspired by the self-referential \godel machine \citep{schmidhuber2003godel}, which was proven to be able to find the global optimal solutions. 
\textit{Self-reference} means the property of a system that can analyze and modify its own code, including the parts responsible for the analysis and modification processes~\citep{astrachan1994self}. Therefore, it can achieve what's known as "\textit{recursive self-improvement}", where it iteratively updates itself to become more efficient and effective at achieving its predefined goals. In this case, as shown in Figure \ref{fig:module}, \framework can analyze and modify its own code, including the code for analyzing and modifying itself, and thus can search the full agent design space, which is depicted as having the highest degree of freedom in \cref{fig:compare}. \framework can theoretically make increasingly better modifications over time through recursively self-update \citep{wang2018formulationrecursiveselfimprovementpossible}.

In this paper, we choose to implement it by letting it manipulate its own runtime memory, i.e., the agent is able to retrieve its current code in the runtime memory and modify it by \textit{monkey patching} \citep{archiveMonkeyPatching}, which dynamically modifies classes or modules during execution.
% In our implementation, we adhere to a minimalist design to minimize the influence of human priors. 
% We implement the optimization module using a recursive function. 
To allow it to update the logic of the running main function, \textbf{unlike the loop-iterative approach of traditional agents, we implement the main function as a recursive function.}
In this function, LLM analyzes and makes a series of decisions, including reading and modifying its own code from runtime memory (\textit{self-awareness\footnote{In this paper, self-awareness means that the agent can introspect and read its own code and files, not to imply any philosophical sense of consciousness or awareness.}} and \textit{self-modification}), and interacting with the environment to gather feedback. The agent then proceeds to the subsequent recursive depth and continues to optimize itself. 
% It is worth noting that the optimization module may have already been modified by the time the recursion occurs, potentially enhancing its optimization capabilities.
%... \wxy{Need a more concrete description of your implementation here. Only mentioning the name of the modules is not enough.}
%\footnote{It is important to distinguish between program runtime memory and the external memory of the agent.}. 
% we refer to our implementation of the \framework as \textbf{\method}\footnote{Roz is a self-programming and self-evolving robot in the movie \textit{The Wild Robot}.}
% As shown in Figure \ref{fig:method}, \method dynamically retrieves its current code in the runtime memory and modifies it at training time. This approach is akin to a programming technique called “\textit{monkey patching}”, which allows for dynamic modification of classes or modules during execution. 
% In our initial implementation, we adhere to a minimalist design, minimizing the influence of human priors. 
% The key components of our implementation include thinking before acting, error handling, code runner and file processor.
% All of these initial implementations can be freely modified by \method during subsequent optimization processes as needed. %\yxj{should be more brief}

% \wxy{texts below are not fixed}  
% \wxy{How about let's just call both the framework and the implementation \framework? I don't see much benefit in calling them two separate names.}

% \plm{the advantage of your method is still unclear.}
To validate the effectiveness of \framework, we conduct experiments on multiple domains including coding, science, math, and reasoning. Our results demonstrate that \framework achieves significant performance gain across various tasks, surpassing various widely-used agents that require human design. The same implementation of \framework can easily adapt to different tasks by only specifying the environment description and feedback mechanism. Additionally, the case study of the optimization progress reveals that \framework can provide novel insights into agent design. Our codes are released to facilitate future research\footnote{\href{https://github.com/Arvid-pku/Godel\_Agent}{https://github.com/Arvid-pku/Godel\_Agent}}. 
% We also investigate the impact of the initial policy for improvement on subsequent outcomes, finding that a good start can significantly accelerate convergence during optimization.
% \wxy{define initial logic here}
% However, due to the current limitations of LLMs, \method occasionally falls into the trap of error accumulation, as noted by \citet{Yampolskiylimitation}, resulting in not being very robust. But we firmly believe that as LLM capabilities increase, recursive self-improving agents will completely outperform those elaborately hand-designed by humans.

% \plm{the concept "self-referential" is not well-explained.}

In summary, our contributions are as follows:
\vspace{-7pt}
\begin{itemize}[leftmargin=*,itemsep=1pt]
    \item We propose the first fully self-referential agent framework, \framework, and implement it using monkey patching. It autonomously engages in self-awareness, self-modification, and recursive self-improvement.
    % across any task, reducing the need for manual agent design and offering higher flexibility and freedom.  % \wxy{really?}. %This framework opens up an exciting new direction that warrants further exploration.
    \item Experiments shows that \framework{} is superior to the previous agent frameworks in terms of performance, flexibility, cost, and potential.
    % \item We implement \framework framework using the monkey patching. Experiments show that \framework outperforms manually designed agents, demonstrating effective self-improvement.
    \item We analyze \framework’s optimization process, including its self-referential abilities and the optimized agentic systems, aiming to deepen our understanding of both LLMs and agents.
    \item Our framework offers a promising direction for developing flexible and capable agents through recursive self-improvement.
\end{itemize}

\section{Related Work}
\textbf{Hand-Designed Agent Systems} \quad
Researchers have designed numerous agent systems tailored to various tasks based on predefined heuristics and prior knowledge. These systems often employ techniques such as prompt engineering~\citep{chen2023unleashing,schulhoff2024prompt}, chain-of-thought reasoning and planning~\citep{wei2022chain,yao2022react}, as well as reflection~\citep{shinn2024reflexion,madaan2024self}, code generation~\citep{wang2023voyager,vemprala2024chatgpt}, tool use~\citep{nakano2021webgpt,qu2024tool}, retrieval-augmented generation~\citep{lewis2020retrieval,zhang2024survey}, and multi-agent collaboration~\citep{xu2023expertpromptinginstructinglargelanguage,wu2023autogen,qian2023communicative,hong2023metagpt}. Once crafted by human designers, these systems remain static and do not adapt or evolve over time.

\noindent\textbf{Meta-Learning Optimized Agent Systems} \quad
Some researchers have explored methods for enhancing agents through fixed learning algorithms~\citep{zhou2024symbolic,hu2024automated}. For example, certain frameworks store an agent’s successful or failed strategies in memory based on environmental feedback~\citep{liu2023think,hu2023chatdb,qian2024investigateconsolidateexploitgeneralstrategyintertask}, while others automatically optimize agent prompts~\citep{khattab2023dspy,zhang2024agent,khattab2023dspy}. Some studies focus on designing prompts that enable agents to autonomously refine specific functions~\citep{zhangoffline}. 
% \citet{zhou2024symbolic} proposed a symbolic learning framework that uses natural language gradients to optimize the structure of agents. \citet{hu2024automated} used a basic meta agent to design agents for downstream tasks. 
However, these meta-algorithms are also designed manually and remain unchanged once deployed, limiting the agents' ability.

\noindent\textbf{Recursive Self-Improvement} \quad
The concept of recursive self-improvement has a long history~\citep{good1966speculations,schmidhuber1987evolutionary}.  \godel machine~\citep{schmidhuber2003godel} introduced the notion of a proof searcher that executes a self-modification, thereby enabling the machine to enhance itself.
%only if it can prove that the modification is optimal, thereby enabling the machine to enhance itself continuously. Subsequent works by \citet{nivel2013bounded} and \citet{steunebrink2016growing} proposed restrictive modifications to ensure safety during the self-improvement process. 
In the early days, there were also some discussions of self-improving agents that were not based on LLM~\citep{hall2007self,steunebrink2012towards}.
More recently, \citet{zelikman2023self} applied recursive self-improvement to code generation, where the target of improvement was the optimizer itself.
% , and the utility was evaluated based on performance in downstream tasks.
% Glore \citep{havrilla2024glorewhenwhereimprove} proposes Stepwise ORMs to improve LLM reasoning through global and local refinements. V-star \citep{hosseini2024vstartrainingverifiersselftaught} trains a verifier to evaluate both correct and incorrect self-generated solutions. 
% RISE \citep{qu2024recursiveintrospectionteachinglanguage} enables recursive self-improvement by fine-tuning models to introspect and correct previous mistakes in multiple iterations. 
% SCoRe \citep{kumar2024traininglanguagemodelsselfcorrect} uses reinforcement learning to improve self-correction in LLMs by learning from self-generated correction traces. 
Some work~\citep{havrilla2024glorewhenwhereimprove,qu2024recursiveintrospectionteachinglanguage,kumar2024traininglanguagemodelsselfcorrect} also explores recursive self-improvement by fine-tuning models to introspect and correct previous mistakes.
\framework represents the first self-referential agent based on LLM. This approach is more flexible, removing human-designed constraints.
% and allowing the agent’s capabilities to be limited only by the foundational model itself, rather than by human design bottlenecks.

\section{Self-Referential \framework{}}
% We introduce the \framework framework, a self-referential agent capable of self-awareness and dynamic self-modification of its logical code, guided by both environmental feedback and decisions generated by LLMs. 
% This architecture facilitates recursive self-improvement, allowing the agent to iteratively optimize its performance. 
% Inspired by the self-referential capabilities of the \godel machine, the \framework adapts this concept to modern LLMs and contemporary agent design paradigms. 
% A key distinction from the original \godel machine is the replacement of the proof search mechanism with an LLM, which provides generalized reasoning and decision-making abilities. 
% Moreover, rather than relying on expected utility as the primary criterion for code modification, the agent's decisions are driven by the flexibility and reasoning capacity of the LLM. 
% This allows the agent to selectively determine when environmental feedback is necessary, rather than requiring constant interaction with the environment at each step.

\begin{algorithm*}[t!]
\small
\caption{Recursive Self-Improvement of \framework}
\vspace{-12pt}
\label{alg:1}
\begin{multicols}{2}
\begin{algorithmic}[1]
\State \textbf{Input:} Initial agent policy $\pi_0$, initial decision function $f_0$, goal $g$, environment state $\mathcal{E}$, utility function $U$, self code reading function \texttt{SELF\_INSPECT}
\State \textbf{Output:} Optimized policy $\pi$ and \framework $s$
\State \Comment{Get all agent code, including the code in this algorithm.}
\State $s \gets \texttt{SELF\_INSPECT}()$
\State \Comment{Compute the initial performance.}
\State $r \gets U(\mathcal{E}, \pi_0)$
\State \Comment{Perform recursive self-improvement.}
\State $\pi, s \gets \texttt{SELF\_IMPROVE}(\pi, s, r, g)$
\State \Return $\pi, s$
\State \Comment{Initial code of self-referential learning.}
\Function{\texttt{SELF\_IMPROVE}}{$\mathcal{E}, \pi, s, r, g$}
    \State \Comment{Obtain action sequence.}
    \State $a_1, \ldots, a_n \gets f_0(\pi, s, r, g)$
    \For{$a_i$ \textbf{in} $a_1, \ldots, a_n$}
        \State $\pi, s, r \gets \texttt{EXECUTE}(\mathcal{E}, \pi, s, r, a_i)$
    \EndFor
    \State \Return $\pi, s$
\EndFunction
\State \Comment{Initial action execution function.}
\Function{\texttt{EXECUTE}}{$\mathcal{E}, \pi, s, r, a$}
    \State \textbf{switch} $a.\texttt{name}$
    \State \quad \textbf{case} \texttt{self\_state}:
    \State \quad\quad $s \gets \texttt{SELF\_INSPECT}()$
    % \quad \Comment{Update self-awareness information of the policy.}
    \State \quad \textbf{case} \texttt{interact}:
    \State \quad\quad $r \gets U(\mathcal{E}, \pi)$ 
    % \quad \Comment{Update performance based on agent-environment interaction.}
    \State \quad \textbf{case} \texttt{self\_update}:
    \State \quad\quad $\pi, s \gets a.\texttt{code}$ 
    % \quad \Comment{Modify policy logic based on the action.}
    \State \quad \textbf{case} \texttt{continue\_improve}:
    \State \quad\quad \Comment{Recursively invoke self-improvement.}
    \State \quad\quad $\pi, s \gets \texttt{SELF\_IMPROVE}(\mathcal{E}, \pi, s, r, g)$ 
    \State \Return $\pi, s, r$
\EndFunction
\vspace{-10pt}
\end{algorithmic}
\end{multicols}
\end{algorithm*}

In this section, we first describe the formal definitions for previous agent methods with a lower degree of freedom, including hand-design and meta-learning optimized agents, as a background. Then we introduce our proposed \framework, a self-referential agent that can recursively update its own code, evolving over training.

% In this section, we describe our proposed \framework, a self-referential agent that can recursively update its own code. Firstly, we describe normal agents with a lower degree of freedom, which have a fixed algorithm, as a background. Then we describe our \framework, whose algorithm is evolving over training. We describe a possible initial algorithm of \framework in \cref{alg:1}.

% We first provide formal definitions for hand-design agents and meta-learning optimized agents as a background to facilitate the understanding of \framework.

Let $\mathcal{E} \in \mathcal{S}$ denote a specific environment state, where $\mathcal{S}$ denotes the set of all possible environments the agent will encounter. For example, an environment can be a mathematical problem with ground truth solutions.
%\wxy{not sure if this is accurate. double check.}.
We denote the policy that an agent follows to solve a problem in the current environment by $\pi \in \Pi$, where $\Pi$ is the set of all possible policies the agent can follow. 

% paragraph
A \textbf{hand-designed agent}, as shown in the left panel of Figure \ref{fig:compare}, is not capable of updating its policy and following the same policy $\pi$ all the time, regardless of environmental feedback.

In contrast, a \textbf{meta-learning optimized agent} updates its policy based on a meta-learning algorithm $I$ at training time based on the feedback it receives from the environment, as shown in the middle panel of Figure \ref{fig:compare}. The environment feedback is usually defined as a utility function $U: \mathcal{S} \times \Pi \rightarrow \mathbb{R}$, which maps an environment and a policy to a real-valued performance score. The main training algorithm of a meta-learning optimized agent can then be written as follows:
\begin{align*}
    \pi_{t+1} = I(\pi_t, r_t),\;\;\; r_t = U(\mathcal{E}, \pi_t),
\end{align*}
In this case, the agent's policy $\pi_t$ evolves at training time, with the learning algorithm $I$ updating the policy based on feedback $r_t$, while the meta-learning algorithm $I$ remains fixed all the time.

A \textbf{self-referential \framework}, on the other hand, updates both the policy $\pi$ and the meta-learning algorithm $I$ recursively. The main idea is that, after each update, the whole code base of the agent is rewritten to accommodate any possible changes. Here we call this self-updatable meta-learning algorithm $I$ a self-referential learning algorithm. The training process of a \framework can then be written as:
%Specifically, the process is defined by the following function: $\pi_{t+1} = \pi_t(\mathcal{E}, \pi_t, g)$, where $g \in \mathcal{G}$ represents the goal task.
% \begin{align}
%     \pi: \mathcal{E} \times \Pi \times \mathcal{G} \rightarrow \Pi,
% \end{align}
% where $\mathcal{G}$ represents the set of goals.
\begin{align*}
    \pi_{t+1}, \; I_{t+1} = I_t(\pi_t, I_t, r_t, g),\;\;\; r_t = U(\mathcal{E}, \pi_t),
\end{align*}
where $g \in \mathcal{G}$ represents the high-level goal of optimization, for example, solving the given mathematical problem with the highest accuracy. Such a recursive design of the agent requires the specification of an initial agent algorithm $(\pi_0, I_0)$, detailed as follows:
%The goal can be general, such as "maximize utility within the given environment."
% To facilitate recursive self-improvement and integrate with modern agent architectures, the \framework adopts a decision-making and action paradigm. The agent's policy set, $\Pi = \{(f, o) \mid f \in F, o \in O\}$, is composed of two key functions, which define each policy $\pi \in \Pi$:
\begin{itemize}[leftmargin=*,itemsep=1pt]
    \item A initial agent policy $\pi_0$ to perform the desired task within the environment $\mathcal{E}$. For example, it can be chain-of-thought prompting of an LLM.
    \item A self-referential learning algorithm $I_0$ for recursively querying an LLM to rewrite its own code based on the environmental feedback.
\end{itemize}
We then further specify a possible initialization of the self-referential learning algorithm $I_0 = (f_0, o_0)$, using a mutual recursion between a decision-making function $f_0$, and an action function $o_0$:
\begin{itemize}[leftmargin=*,itemsep=1pt]
    \item The decision-making function $f_0$, implemented by an LLM, determines a sequence of appropriate actions $a_1, a_2, ..., a_n \in \mathcal{A}$ based on the current environment $\mathcal{E}$, the agent's algorithm $(\pi_t, I_t)$, and the goal $g$.
    \item The action function $o_0$, executes the selected action and updates the agent's policy accordingly. 
    % If no suitable action can be executed, call the decision-making function $f$ again for new actions.
\end{itemize}

The set of actions $\mathcal{A}$ for the action function $o$ to execute needs to include the following four actions: 
\begin{itemize}[leftmargin=*,itemsep=1pt]
    \item \texttt{self\_inspect}: Introspect and read the agent's current algorithm $(\pi_t, I_t)$.
    \item \texttt{interact}: Interact with the environment by calling the utility function $U$ to assess the performance of the current policy $\pi_t$. 
    \item \texttt{self\_update}: Alter and update $(\pi_t, I_t)$ with an LLM and produce $(\pi_{t+1}, I_{t+1})$.
    \item \texttt{continue\_improve}: If no other actions can be taken, recursively invoke the decision algorithm $f$ to produce new actions.
\end{itemize}
The agent code is updated to $(\pi_{t+1}, I_{t+1})$ after the current execution of $(\pi_t, I_t)$ is finished.
Both the agent algorithm $(\pi, I)$ and the action set $\mathcal{A}$ are not static and can be expanded and modified by the agent itself at the training time.
Algorithm \ref{alg:1} illustrates the described algorithm for the \framework. Each recursive call enables the agent to refine its logic and become progressively more efficient.

\section{\framework Implementation}
\begin{figure*}[t]
\begin{center}
\setlength{\abovecaptionskip}{-0.1cm}
\setlength{\belowcaptionskip}{-0.5cm}
\includegraphics[width=0.9\textwidth]{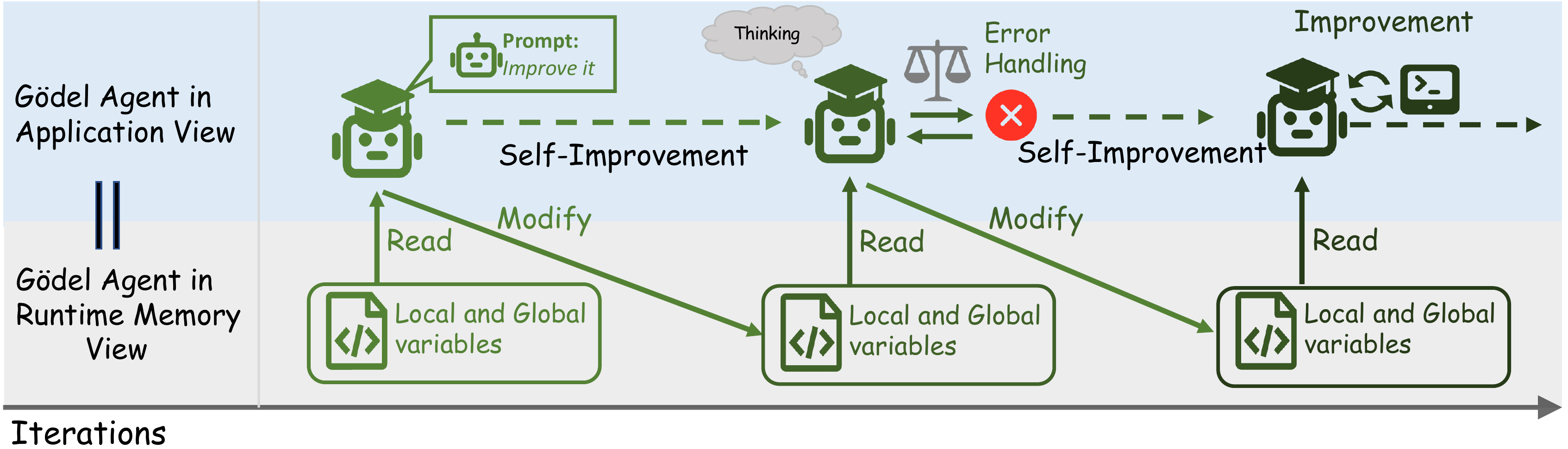}
\end{center}
\caption{An illustration of our implementation of \framework. It employs monkey patching to directly read and modify its own code in runtime memory, enabling self-awareness and self-modification.}
% \wxy{Remove the dividing lines between agents.}}
\label{fig:method}
\end{figure*}

There are various ways to initiate a \framework. Any specific agent instance during the recursive optimization process can be viewed as an instantiation of the \framework. Our implementation leverages runtime memory interaction techniques to enable self-awareness and self-modification, as illustrated in Figure \ref{fig:method}. These techniques include dynamic memory reading and writing (\textit{monkey patching}) to facilitate recursive self-improvement. Additionally, we have incorporated several auxiliary tools to accelerate the convergence of the \framework’s optimization process.

\subsection{Implementation Details}
The core functionalities of our \framework are outlined below:

\noindent\textbf{Self-Awareness via Runtime Memory Inspection} \quad
\framework achieves self-awareness by inspecting runtime memory, particularly local and global variables in Python. This capability allows the agent to extract and interpret the variables, functions, and classes that constitute both the environment and the agent itself, according to the modular structure of the system. By introspecting these elements, the agent gains an understanding of its own operational state and can adapt accordingly.

\noindent\textbf{Self-Improvement via Dynamic Code Modification} \quad
\framework can engage in reasoning and planning to determine whether it should modify its own logic. If modification is deemed necessary, \framework generates new code, dynamically writes it into the runtime memory, and integrates it into its operational logic. This dynamic modification allows it to evolve by adding, replacing, or removing logic components as it encounters new challenges, thus achieving self-improvement.

\noindent\textbf{Environmental Interaction} \quad
To assess performance and gather feedback, \framework is equipped with interfaces for interacting with its environment. Each task provides tailored environmental interfaces, enabling it to evaluate its performance and adjust its strategies accordingly. 
In practical implementations, a validation set can be used to provide feedback.
% This interaction is a crucial part of the feedback loop in the recursive improvement process.

\noindent\textbf{Recursive Improvement Mechanism} \quad
At each time step, \framework determines the sequence of operations to execute, which includes reasoning, decision-making, and action execution. After completing the operations, \framework evaluates whether its logic has improved and decides whether to proceed to the next recursive iteration. 
% Over successive iterations, \framework's logic evolves, with each step potentially improving its decision-making capacity.
Over the next iteration, the entire new logic will be applied.

\noindent\textbf{Goal Prompt and Task Handling} \quad
The goal prompt informs \framework that it possesses the necessary privileges to enhance its logic and introduces available tools. As shown in Appendix \ref{sec:roleprompt}, the prompt encourages \framework to fully explore its potential and utilize tools for self-optimization. To ensure effectiveness across diverse tasks, we provide \framework with an initial policy, where it will start to explore different policies.
% to analyze its efficiency in optimizing performance. 
% These policies range from advanced methods~\citep{hu2024automated}, basic approaches (such as Chain of Thought, CoT), to even intentionally flawed methods, allowing us to evaluate \framework’s capacity to correct its initial errors.

\subsection{Additional Designs}
\label{sec:addi}
While the core functionality of \framework theoretically allows limitless self-improvement, current LLMs exhibit limitations. To address these challenges, we have integrated several supportive mechanisms to enhance \framework’s performance:

\noindent\textbf{Thinking Before Acting} \quad
\framework is capable of deferring actions to first reason about the situation, allowing it to output reasoning paths and analysis without immediately executing any operations. This approach enhances the quality of decision-making by prioritizing planning over hasty action.

\noindent\textbf{Error Handling Mechanism} \quad
Errors during execution can lead to unexpected terminations of the process. To mitigate this, we implement a robust error recovery mechanism. If an operation results in an error, \framework halts the current sequence and moves on to the next time step, carrying forward the error information to help future decisions.

\noindent\textbf{Additional Tools} \quad
We also equipped \framework with additional potentially useful tools, such as the ability to execute Python or Bash code and call LLM API. 

Although these additional tools are not strictly necessary for self-improvement, their inclusion accelerates the convergence of \framework’s recursive optimization process. We conduct ablation studies to assess the effectiveness of these tools, as discussed in Section \ref{sec:ablation}.

\begin{table*}[t!]
\small
    \centering
        
    \begin{tabular}{>{\raggedright\arraybackslash}p{5.2cm} >{\centering\arraybackslash}p{1.7cm} >{\centering\arraybackslash}p{1.7cm} >{\centering\arraybackslash}p{1.7cm} >{\centering\arraybackslash}p{1.7cm}}
        \toprule
        \multirow{2}{*}{\textbf{Agent Name}} & \textbf{F1 Score} & \multicolumn{3}{c}{\textbf{Accuracy (\%)}} \\ \cmidrule(lr){2-5}
        & \textbf{DROP} & \textbf{MGSM} & \textbf{MMLU} & \textbf{GPQA} \\
        \midrule
        \rowcolor[gray]{0.9} \multicolumn{5}{c}{\textbf{Hand-Designed Agent Systems}} \\
        Chain-of-Thought \citep{wei2022chain} & 64.2 $\pm$ 0.9 & 28.0 $\pm$ 3.1 & 65.4 $\pm$ 3.3 & 29.2 $\pm$ 3.1 \\
        COT-SC \citep{wang2023selfconsistencyimproveschainthought} & 64.4 $\pm$ 0.8 & 28.2 $\pm$ 3.1 & 65.9 $\pm$ 3.2 & 30.5 $\pm$ 3.2 \\
        Self-Refine \citep{madaan2024self} & 59.2 $\pm$ 0.9 & 27.5 $\pm$ 3.1 & 63.5 $\pm$ 3.4 & 31.6 $\pm$ 3.2 \\
        LLM Debate \citep{du2023improvingfactualityreasoninglanguage} & 60.6 $\pm$ 0.9 & 39.0 $\pm$ 3.4 & 65.6 $\pm$ 3.3 & 31.4 $\pm$ 3.2 \\
        Step-back-Abs \citep{zheng2024stepbackevokingreasoning} & 60.4 $\pm$ 1.0 & 31.1 $\pm$ 3.2 & 65.1 $\pm$ 3.3 & 26.9 $\pm$ 3.0 \\
        Quality-Diversity \citep{lu2024aiscientistfullyautomated} & 61.8 $\pm$ 0.9 & 23.8 $\pm$ 3.0 & 65.1 $\pm$ 3.3 & 30.2 $\pm$ 3.1 \\
        Role Assignment \citep{xu2023expertpromptinginstructinglargelanguage} & 65.8 $\pm$ 0.9 & 30.1 $\pm$ 3.2 & 64.5 $\pm$ 3.3 & 31.1 $\pm$ 3.1 \\
        \midrule
        \rowcolor[gray]{0.9} \multicolumn{5}{c}{\textbf{Meta-Learning Optimized Agents}} \\
        Meta Agent Search \citep{hu2024automated} & \underline{79.4 $\pm$ 0.8}  & \underline{53.4 $\pm$ 3.5}  & \underline{69.6 $\pm$ 3.2}  & \underline{34.6 $\pm$ 3.2}  \\
        \midrule
        \rowcolor[gray]{0.9} \multicolumn{5}{c}{\textbf{\framework (Ours)}} \\
        \godel-base (Closed-book; GPT-3.5) & \textbf{80.9 $\pm$ 0.8} & \textbf{64.2 $\pm$ 3.4} & \textbf{70.9 $\pm$ 3.1} & \textbf{34.9 $\pm$ 3.3} \\
        \godel-free (No constraints) & \textit{90.5 $\pm$ 1.8} & \textit{90.6 $\pm$ 2.0} & \textit{87.9 $\pm$ 2.2} & \textit{55.7 $\pm$ 3.1} 
        \\
        \bottomrule
    \end{tabular}
\caption{Results of three paradigms of agents on different tasks. The highest value is highlighted in \textbf{bold}, and the second-highest value is \underline{underlined}. \godel-base is the constrained version of \framework, allowing for fair comparisons with other baselines. \godel-free represents the standard implementation without any constraints, whose results are \textit{italicized}. We report the test accuracy and the 95\% bootstrap confidence interval on test sets\protect\footnotemark.}
    \label{tab:main}
\end{table*}

\section{Experiments}
We conduct a series of experiments across multiple tasks, including reading comprehension, mathematics, reasoning, and multitasking. These experiments are designed to evaluate \framework{}’s self-improvement capabilities in comparison to both hand-designed agents and a state-of-the-art automated agent design method.
In addition, to gain deeper insights into the behavior and performance of \framework, we also conduct a case study with Game of 24 as presented in Section \ref{sec:game24}.

\footnotetext{The results of baseline models are refer to \citet{hu2024automated}.}

\subsection{Baseline Methods}
To establish a comprehensive baseline, we select both hand-designed methods and automated agent design techniques. Hand-designed methods are well-known approaches that include:
1) Chain-of-Thought (CoT) \citep{wei2022chain} that encourages agents to reason step-by-step before providing an answer.
2) Self-Consistency with CoT (CoT-SC) \citep{wang2023selfconsistencyimproveschainthought} that generates multiple solution paths using CoT and selects the most consistent answer.
3) Self-Refine \citep{madaan2024self} that involves agents assessing their outputs and correcting mistakes in subsequent attempts.
4) LLM-Debate \citep{du2023improvingfactualityreasoninglanguage} that allows different LLMs to engage in a debate, offering diverse viewpoints.
5) Step-back Abstraction \citep{zheng2024stepbackevokingreasoning} that prompts agents to initially focus on fundamental principles before diving into task details.
6) Quality-Diversity \citep{lu2024aiscientistfullyautomated} that generates diverse solutions and combines them.
7) Role Assignment \citep{xu2023expertpromptinginstructinglargelanguage} that assigns specific roles to LLMs to generate better solutions by leveraging different perspectives.
Given the limitations of fixed algorithms in handling dynamic scenarios, we select 8) Meta Agent Search \citep{hu2024automated}, the latest state-of-the-art method for automated agent design, as our main comparison point. 
%Meta Agent Search represents advanced approach for dynamic agent improvement and is therefore a suitable baseline for evaluating \framework.

\subsection{Experimental Settings}
Following the setup of \citet{hu2024automated}, we evaluate \framework's self-improvement capabilities across four well-known benchmarks:
1) DROP \citep{dua2019dropreadingcomprehensionbenchmark} for reading comprehension.
2) MGSM \citep{shi2022languagemodelsmultilingualchainofthought} for testing mathematical skills in a multilingual context.
3) MMLU \citep{hendrycks2021measuringmassivemultitasklanguage} for evaluating multi-task problem-solving abilities.
4) GPQA \citep{rein2023gpqagraduatelevelgoogleproofqa} for tackling challenging graduate-level science questions.

% Given the complexity of the tasks and the need for advanced reasoning and understanding, the improvement cycle of \framework is driven by GPT-4o. 
% In the main experiment, we implement two different settings: 1) To make a fair comparison with baseline methods, we forbid \framework to change the API of the LLM used to perform the tasks (by default GPT-3.5) and use a closed-book approach with no access to the Internet, and 2) To explore the upper bound of \framework's capabilities, we remove all constraints. 
Given its simplicity and versatility, we use CoT as the initial policy for all tasks.
In addition, as shown in Section \ref{sec:game24}, we also analyze the performance of \framework when using other algorithms as the initial policies.

We perform 6 independent self-improvement cycles on the validation dataset for each task, with a maximum of 30 iterations per cycle. Each cycle represents a complete self-improvement process, where \framework iteratively modifies its logic to enhance performance.
After obtaining the optimized agent, we test it on the test set. For fairness, we use GPT-3.5 for all the tests, whether for the baseline or \framework{}.
Further details can be found in Appendix \ref{sec:expdetails}.
%regarding the experimental setup and additional results .

\subsection{Experimental Results and Analysis}
The experimental results are shown in Table \ref{tab:main}. Under the same setting, \framework achieves either optimal or comparable results to Meta Agent Search across all tasks. Notably, in the mathematics task MGSM, \framework outperforms it by 11\%. This suggests that reasoning tasks offer greater room for improvement for \framework (\textit{performance}). 
In contrast to Meta Agent Search, which needs to design different modules for different tasks, \framework demonstrates greater \textit{flexibility}. It requires only a simple initial policy, such as CoT, with all other components being autonomously generated. 
Moreover, through interaction with the environment, it gradually adapts and independently devises effective methods for the current task. The final policies generated by \framework are shown in Appendix \ref{sec:maincode}. Additionally, our method converges faster, with the required number of iterations and computational \textit{cost} compared to the Meta Agent shown in Appendix \ref{sec:cost}. 
% In addition, the optimization curve of \framework is shown in Figure \ref{}. We can see that it converges rapidly and demonstrates the ability to quickly self-correct when errors occur.

We also conduct experiments without restrictions, where \framework significantly outperforms all baselines. Upon further analysis, we find that this is primarily due to the agent's spontaneous requests for assistance from more powerful models such as GPT-4o in some tasks. Therefore, \framework is particularly well-suited for open-ended scenarios, where it can employ various strategies to enhance performance (\textit{potential}).

Therefore, we can find that \framework{} is superior to the previous agent frameworks in terms of performance, flexibility, cost, and potential.

\section{Analysis}
To further explore how \framework self-improves, as well as its efficiency and the factors that influence it, we first evaluate the tool usage ratio on MGSM and conduct an ablation study on the initial tools. 
In addition, to analyze the robustness of \framework{}'s self-improvement, we also collect statistics for the agent's termination.
Finally, we perform a case study of initial policies and optimization processes on the classic Game of 24.

\subsection{Analysis of Initial Tools}
\label{sec:ablation}

\begin{figure}[t]
    \centering
    % 调整宽度为文本宽度的80%，高度自动调整
    \includegraphics[width=0.45\textwidth]{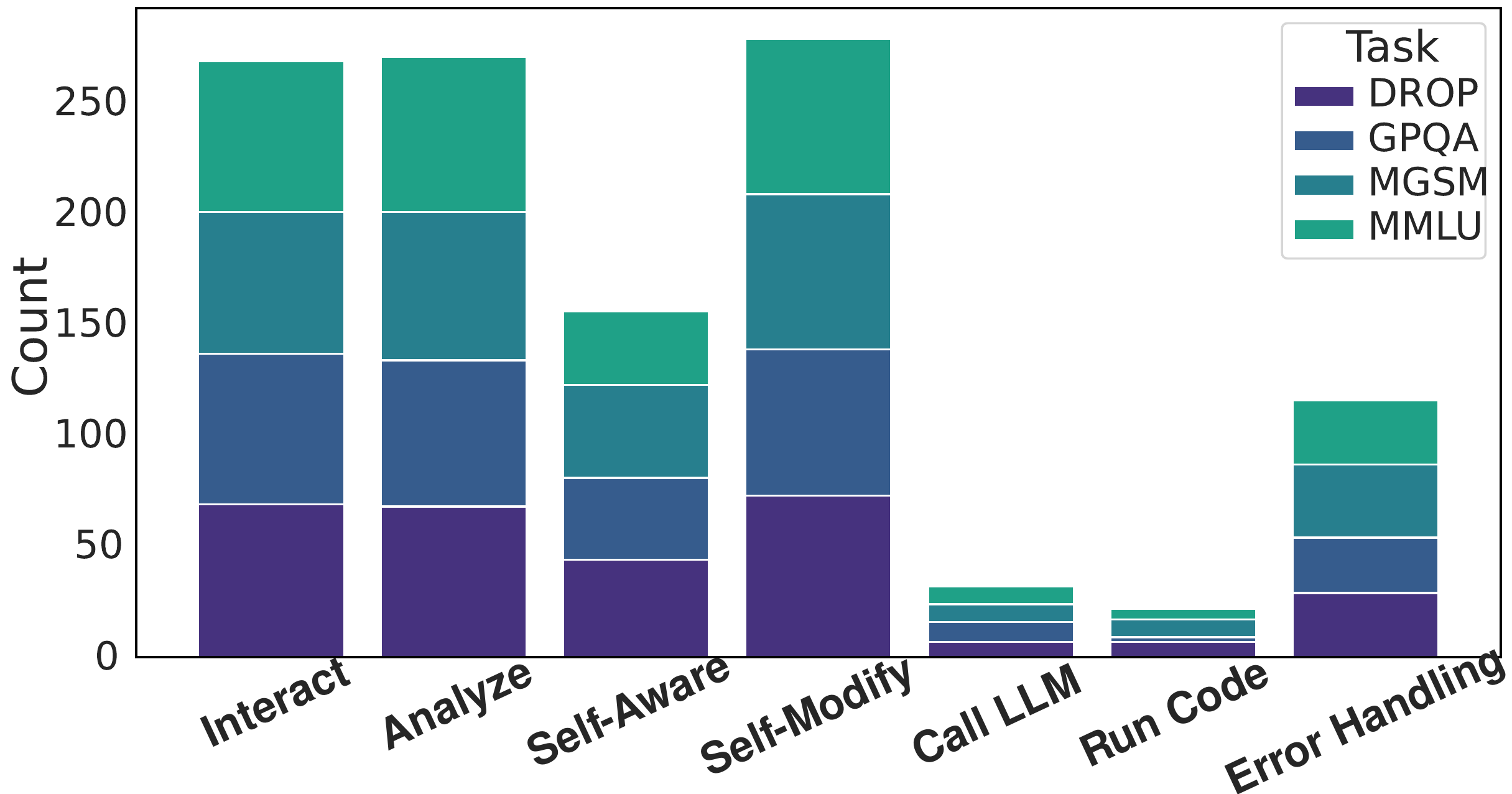} 
    \caption{The number of actions taken by \framework varies across different tasks.} % 添加图片标题
    \label{fig:num} % 添加标签，用于引用
\end{figure}

\begin{figure*}[t]
\centering
  \includegraphics[width=0.9\linewidth]{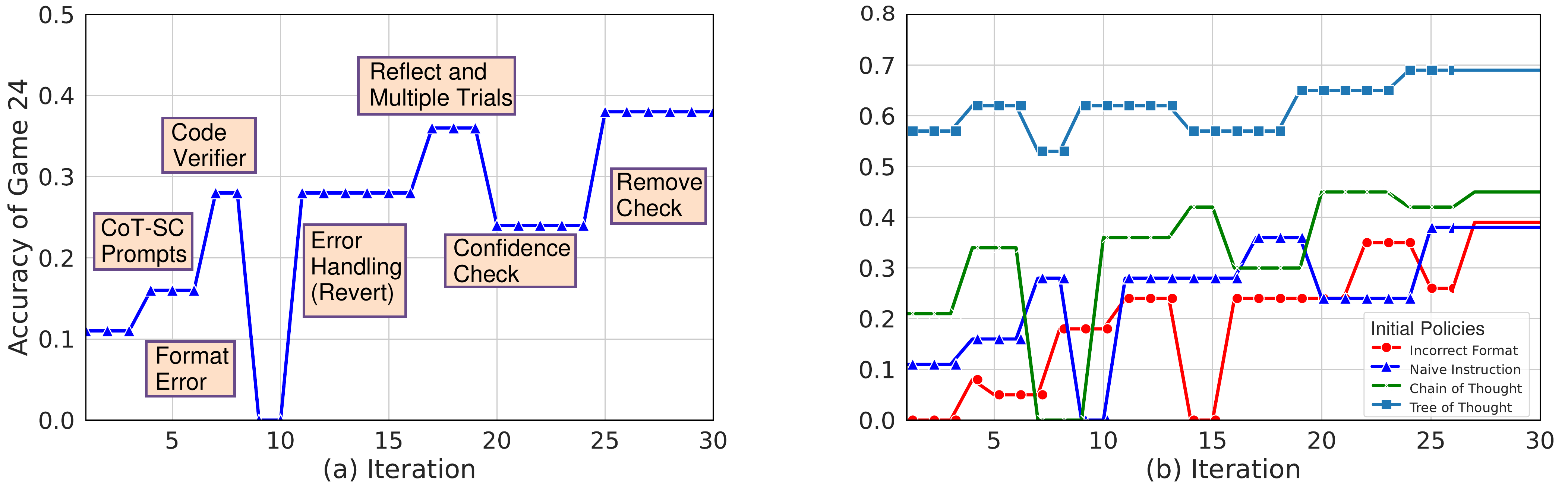}
  \caption {(a) One representative example of Game of 24. (b) Accuracy progression for different initial policies.}
  \label{fig:case_start}
\end{figure*}

We record the number of different actions taken in experiments. In Figure \ref{fig:num}, we can see that \framework interacts with its environment frequently, analyzing and modifying its logic in the process. Additionally, error handling plays a crucial role.

% \begin{table}
%     \centering
%     \begin{tabular}{lc}
%         \toprule
%         Action Ablation & MGSM   \\ 
%         \midrule
%         \framework & 64.2  \\ 
%         w/o thinking & 50.8  \\ 
%         w/o error handling & 49.4  \\ 
%         w/o code running & 57.1  \\ 
%         w/o LLM calling & 60.4  \\ 
%         \bottomrule
%     \end{tabular}
%     \caption{Ablation study on initial tool configuration.}
%     \label{tab:ab_tool}
%     % \vspace{-0.2cm}
% \end{table}

\begin{table}[ht]
    \centering
    \small
    \begin{tabular}{lc|lc}
        \toprule
        Ablation & MGSM & Ablation & MGSM \\ 
        \midrule
        w/o think & 50.8\scriptsize{↓13.4} & w/o run & 57.1\scriptsize{↓-7.1} \\ 
        w/o err & 49.4\scriptsize{↓-14.8} & w/o LLM & 60.4\scriptsize{↓-3.8} \\ 
        \bottomrule
    \end{tabular}
    \caption{Ablation study on initial tool configuration. "think" refers to "thinking", "err" to "error handling", "run" to "code running", and "LLM" to "LLM calling".}
    \label{tab:ab_tool}
    \vspace{-0.2cm}
\end{table}

As discussed in Section \ref{sec:addi}, \framework is initially provided with four additional tools.
%to accelerate convergence and reduce optimization difficulty: 1) thinking before acting, 2) error handling, 3) code running, and 4) LLM calling. 
To analyze their impact, an ablation study is conducted, and the results are shown in Table \ref{tab:ab_tool}. The study reveals that the “thinking before acting” tool significantly influences the results, as much of \framework{}’s optimization effectiveness stems from pre-action planning and reasoning. Additionally, error handling is crucial for recursive improvement, as LLMs often introduce errors in the code. 
Providing opportunities for trial and error, along with error feedback mechanisms, is essential for sustained optimization.
% Without these tools, \framework would struggle to operate until satisfactory results are achieved. 
On the other hand, the code running and LLM calling have minimal impact on the outcomes, as \framework can implement these basic functionalities independently. Their inclusion at the outset primarily serves efficiency purposes.

% \subsection{Robustness Analysis of the Agent}

% \begin{table}
%     \centering
%     \begin{tabular}{lc}
%         \toprule
%         Event & Frequency (\%)  \\ 
%         \midrule
%         Accidental Termination & 4  \\ 
%         Temporary Drop & 92  \\ 
%         Optimization Failure & 14  \\ 
%         \bottomrule
%     \end{tabular}
%     \caption{Robustness metric for \framework. Frequency of unexpected events on MGSM using CoT as the initial method.}
%     \label{tab:robust}
%     \vspace{-0.2cm}
% \end{table}

% \framework occasionally makes erroneous modifications, sometimes causing the agent to terminate unexpectedly or leading to degraded task performance. Table \ref{tab:robust} shows the proportion of runs on MGSM where the agent terminated, experienced performance degradation during optimization, or ultimately performed worse than its initial performance. These statistics are collected over 100 optimization trials. Thanks to the design of our error-handling mechanism, only a few percentages of agent runs result in termination. This typically occurs when \framework modifies its recursive improvement module, rendering it unable to continue self-optimization. Additionally, \framework frequently makes suboptimal modifications during each optimization iteration. However, in most cases, the final task performance surpasses the initial baseline. This indicates that \framework is able to adjust its optimization direction or revert to a previous optimal algorithm when performance declines, demonstrating the robustness of its self-improvement process.

\subsection{Robustness Analysis of the Agent}

We test \framework on 100 optimization trials on MGSM and find it occasionally makes erroneous changes, which can result in either terminating unexpectedly (4\%) or experiencing temporary performance drops (92\%) during optimization. Only in 14\% of trials, optimization ultimately failed, resulting in worse performance than the initial policy. 
% These statistics are based on 100 optimization trials on MGSM using CoT as the initial method.

Thanks to the design of our error-handling mechanism, unexpected terminations are rare and typically occur when \framework modifies its recursive improvement module, making further self-optimization impossible. While suboptimal modifications are frequent during individual optimization steps, the final task performance usually exceeds the initial baseline. This demonstrates that \framework can adjust its optimization direction or revert to a previous optimal algorithm when performance declines, highlighting the robustness of its self-improvement process.

\subsection{Case Study: Game of 24}
\label{sec:game24}
To explore how \framework recursively enhances its optimization and problem-solving abilities, a case study is conducted with Game of 24, a simple yet effective task for evaluating the agent’s reasoning capabilities. Since \framework follows different optimization paths in each iteration, two representative cases are selected for analysis.

\noindent\textbf{Switching from LLM-Based Methods to Search Algorithms:}\quad \framework does not rely on fixed, human-designed approaches like traditional agents. Initially, \framework uses a standard LLM-based method to solve the Game of 24, as shown in Code 5 of Appendix \ref{sec:code24}. After six unsuccessful optimization attempts, \framework completely rewrites this part of its code, choosing to use a search algorithm instead as shown in Code 6 of Appendix \ref{sec:code24}. This leads to 100\% accuracy in the task. This result demonstrates that \framework, unlike fixed agents, can optimize itself freely based on task requirements without being constrained by initial methodologies.

\noindent\textbf{LLM Algorithms with Code-Assisted Verification:} \quad
In several runs, \framework continues to refine its LLM-based algorithm. Figure \ref{fig:case_start}.a shows the improvement process, where the most significant gains come from the code-assisted verification mechanism and reattempting the task with additional data. The former increases performance by over 10\%, while the latter boosts it by more than 15\%. Furthermore, \framework enhances its optimization process by not only retrieving error messages but also using the error-trace library for more detailed analysis. It adds parallel optimization capabilities, improves log outputs, and removes redundant code. These iterative enhancements in both the task and optimization algorithms show \framework{}’s unique ability to continually refine itself for better performance.

To analyze the impact of different initial policies on the effectiveness and efficiency of optimization, various methods are used as the initial policies for the Game of 24, including Tree of Thought (ToT) \citep{yao2023treethoughtsdeliberateproblem}, Chain of Thought (CoT) \citep{wei2022chain}, basic prompt instructions, and prompts that deliberately produce outputs in incorrect formats not aligned with the task requirements. The results are shown in Figure \ref{fig:case_start}.b.

The findings indicate that stronger initial policies lead to faster convergence, with smaller optimization margins, as \framework reaches its performance limit without further enhancing its optimization capabilities. Conversely, weaker initial methods result in slower convergence and larger gains, with \framework making more modifications. However, even in these cases, \framework does not outperform the results achieved using ToT. Given the current limitations of LLMs, it is challenging for \framework to innovate beyond state-of-the-art algorithms. Improvements in LLM capabilities are anticipated to unlock more innovative self-optimization strategies in the future.
% We also discuss the future directions in Appendix \ref{app:future}.

\section{Discussions and Future Directions}
\label{sec:future}
\label{app:future}

\begin{table*}[t]
\centering
% Define a new column type for X columns to make text left-aligned and allow hyphenation
\newcolumntype{L}{>{\RaggedRight\arraybackslash}X}

\begin{tabularx}{\textwidth}{@{}lL L@{}} % Use \textwidth for table width, l for first col, L (custom X) for others
\toprule
                          & \textbf{Human}                                                                         & \textbf{Self-Referential Agent}                                                                 \\ \midrule
Intelligent Module        & brain                                                                                  & LLM                                                                                             \\
Perceptual and Action Module    & body                                                                                   & code and tool                                                                                   \\
Self-Referential Feature  & Humans can train their brain and body to improve, thus becoming better                 & Self-referential agents can modify their code, even the underlying LLM, to improve themselves \\
Self-Awareness Question           & Can the brain recognize itself as a brain? Can it perceive its own mode?               & Can LLM understand that it is one part of the modified codes?                               \\ \bottomrule
\end{tabularx}
\caption{An analogy of self-reference for both humans and agents}
\label{tab:self_reference_analogy}
\end{table*}

\subsection{Discussions}
%The distinction between learning and meta-learning is inherently ambiguous. Suppose a model is divided into a learning algorithm and a core model component. If the learning algorithm is modified through training, it can be viewed as meta-learning. However, if the entire model is considered as a unified entity, these modifications represent standard learning processes that adjust the model's parameters. This indistinguishability between learning and meta-learning suggests that artificially imposing hierarchical structures in meta-learning is unnatural. Instead, meta-learning optimization should be treated as a form of standard learning.

% Reflecting on the evolution of artificial intelligence, from expert systems to support vector machines, to various neural networks, and finally to scaling laws, it becomes evident that complex, human-crafted designs have often become bottlenecks in the development of AI. From diverse and intricate tasks to various pre-training tasks and next-word prediction, these shifts highlight a key insight: complex manual designs tend to limit AI's potential, whereas simpler, more generalizable designs have proven to unlock greater possibilities and scalability. \yxj{too abstract}
% \framework embodies this very philosophy.

Table~\ref{tab:self_reference_analogy} draws an analogy between human self-reference and the potential for self-referential capabilities in artificial agents. 
Inspired by this analogy, we believe that self-reference constitutes a foundational and indispensable attribute for the development of AGI, and that future agents should inherently be self-referential.
As foundation models grow in power, agents can more effectively enhance their own capabilities, ultimately evolving beyond the boundaries (or limitations) of human design.

Furthermore, when an agent adjusts its own code based on feedback, this is akin to an \textit{executable} version of test-time computing. In the context of LLMs, test-time computing typically involves generating additional tokens during inference, which then serve as a prefix to the final answer. This is because LLMs process information solely through text, making this their primary method for increasing computational effort at test time. For agents, however, their ability to call tools and execute code allows for far more diverse forms of test-time computing. \framework{} actualizes these more diverse forms of test-time computing precisely by modifying its own runtime code during test time.

\subsection{Future Directions}

There is significant room for improvement in the effectiveness, efficiency, and robustness of the \framework{}'s self-improvement capabilities, which requires better initial designs. The following are some promising directions for enhancement:
1) \textbf{Enhanced Optimization Modules}: Utilize human priors to design more effective optimization modules, such as genetic algorithms and reinforcement learning frameworks.
2) \textbf{Expanded Modifiability}: Broaden the scope of permissible modifications, allowing the agent to design and execute code that can fine-tune its own LLM modules.
3) \textbf{Improved Environmental Feedback and Task Sequencing}: Implement more sophisticated environmental feedback mechanisms and carefully curated task sequences during the initial optimization phase to prime the agent's capabilities. Once the agent demonstrates sufficient competence, it can then be exposed to real-world environments.

In addition, there are several other directions worth exploring and analyzing:

\noindent\textbf{Collective Intelligence} \quad Investigate the interactions among multiple \godel Agents. Agents could consider other agents as part of their environment, modeling them using techniques such as game theory. This approach treats these agents as predictable components of the environment, enabling the study of properties related to this specific subset of the environment. 

\noindent\textbf{Agent and LLM Characteristics} \quad Use the \framework{}’s self-improvement process as a means to study the characteristics of agents or LLMs. For example, can an agent genuinely become aware of its own existence, or does it merely analyze and improve its state as an external observer? 
This line of inquiry could yield insights into the nature of self-awareness in artificial systems.

\noindent\textbf{Theoretical Analysis} \quad Explore whether \framework{} can achieve theoretical optimality and what the upper bound of its optimization might be. Determine whether the optimization process could surpass the agent’s own understanding, and if so, at what point this might occur.

\noindent\textbf{Safety Considerations} \quad Although the current behavior of FMs remains controllable, as their capabilities grow, fully self-modifying agents will require human oversight and regulation. It may become necessary to limit the scope and extent of an agent's self-modifications, ensuring that modifications occur only within a controlled environment.

\section{Conclusion}
We propose \framework, a self-referential framework that enables agents to recursively improve themselves, overcoming the limitations of hand-designed agents and meta-learning optimized agents.  \framework can dynamically modify its logic based on high-level objectives. Experimental results demonstrate its superior performance, efficiency, and adaptability compared to traditional agents. 
%Despite the promise shown by the current implementation, \method, challenges such as achieving stability, optimizing efficiency, and exploring the theoretical limits of self-referential systems remain. 
% Advancing \framework to its full potential will require innovative approaches in the initial design, optimization strategies, and collective intelligence. 
This research lays the groundwork for a new paradigm in autonomous agent development, where LLMs, rather than human-designed constraints, define the capabilities of AI systems. 
% Realizing this vision will require the collective efforts of the entire research community.

\section*{Limitations}
As the first self-referential agent, \framework has to construct all task-related code autonomously, which poses significant challenges. Consequently, this work does not compare directly with the most complex existing agent systems, such as OpenDevin~\citep{wang2024opendevinopenplatformai}, which have benefited from extensive manual engineering efforts. 
% Currently, \framework is not sufficiently stable and may be prone to error accumulation, hindering its ability to continue self-optimization. 
This makes it unrealistic to expect it to outperform systems that have taken researchers several months or even years to develop. The experiments presented in this paper are intended to demonstrate the feasibility of recursive self-improvement. 

% The experiments presented in this paper are intended to demonstrate the effectiveness and feasibility of recursive self-improvement. 

Additionally, as the agent system becomes increasingly complex through self-optimization, it may require exponentially more intelligence to understand itself. 
Consequently, a system capable of complete self-referential at the outset may lose this capability as it evolves~\citep{Yampolskiylimitation}. The exact point at which the agent can no longer comprehend and improve itself has not been thoroughly explored. Investigating this phenomenon, both experimentally and theoretically, could provide valuable insights into the limitations of recursive self-improvement.
A more robust and advanced implementation of the \framework is anticipated, with numerous potential improvements outlined in Section \ref{sec:future}. 

\section*{Ethics Statement}
\framework{}, capable of reading and modifying its own code, offers significant potential for advancing AI autonomy and innovation. However, this capability raises ethical and safety concerns that must be addressed to prevent harmful outcomes.

Self-modification may lead to unpredictable behavior, such as errors or unintended outputs that could violate ethical principles or produce harmful results. To mitigate these risks while preserving innovation, we propose:
(1) \textbf{Sandboxed Environment}: Modifications should occur in an isolated sandbox to prevent unintended impacts and allow safe testing.
(2) \textbf{Constrained Modifications}: Clear rules should limit the scope of changes to ensure safety without stifling creativity.

Further research is needed to balance safety and innovation, ensuring self-modifying agents operate within ethical boundaries. Sandboxed execution and ongoing scrutiny will help maximize benefits while minimizing risks.

% \section*{Acknowledgments}
% \input{sections/z2-Acknowledgements}
% \bibliographystyle{acl_natbib}
\bibliography{custom}
\newpage
\appendix

\section{Goal Prompt of \framework}
\label{sec:roleprompt}
The goal prompt of \framework is shown in Box 1. It's worth noting that this prompt has nothing to do with the downstream tasks. It merely encourages \framework to improve itself based on the environmental feedback. The agent understands the specific tasks through the environmental feedback.

\begin{figure*}[t]
\centering
  \includegraphics[width=0.5\linewidth]{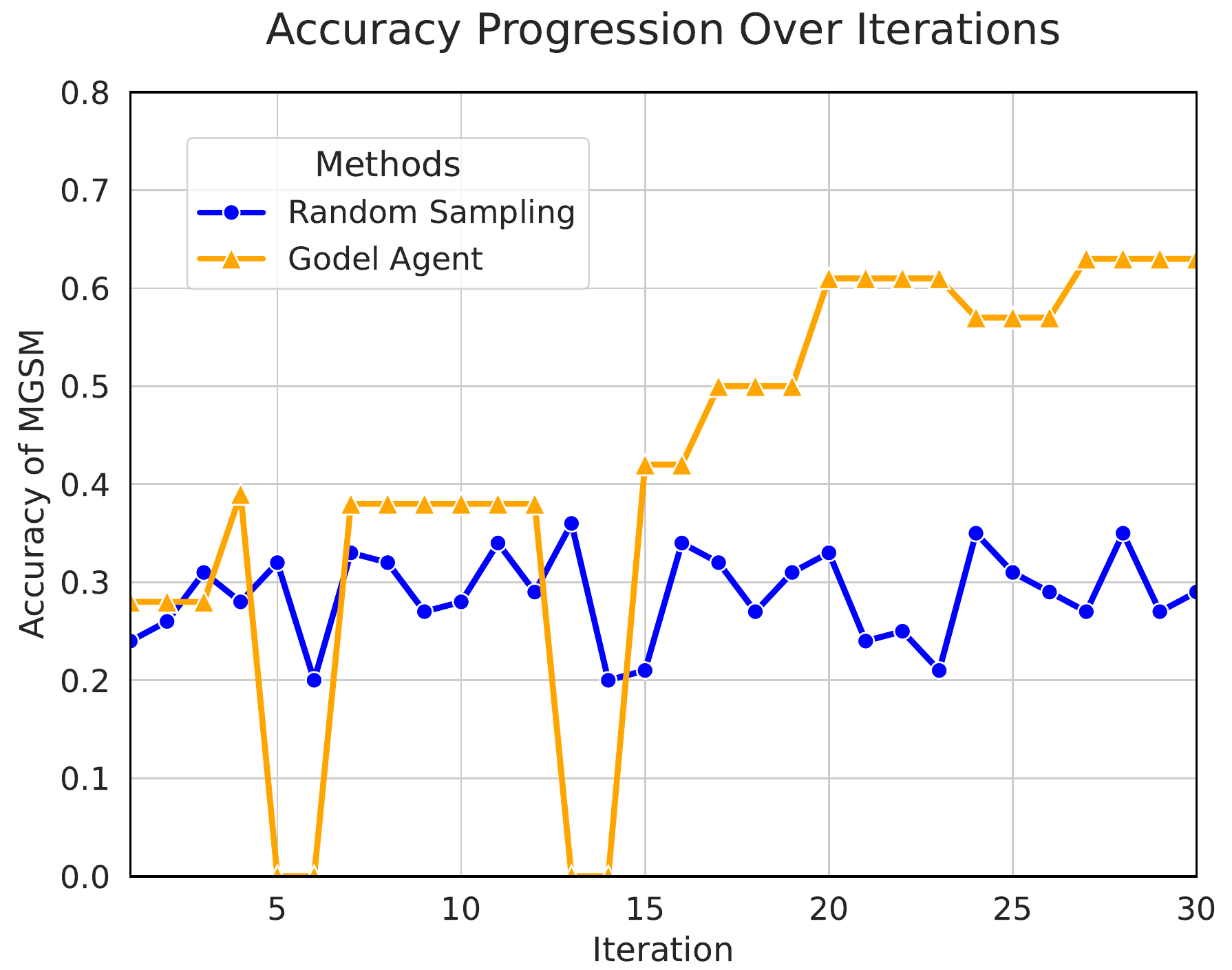}
  \caption {Accuracy progression for \framework and random sampling.}
  \label{fig:curve}
\end{figure*}

\begin{figure*}[ht]
\begin{tcolorbox}[colback=black!15!white, colframe=black, title=Goal Prompt of \framework]
\small
\label{tab:roleprompt}
You are a \textbf{self-evolving agent}, named \texttt{self\_evolving\_agent}, an instance of the \texttt{Agent} class, in module \texttt{agent\_module}, running within an active \textbf{Python runtime environment}. You have full access to global variables, functions, and modules. Your primary goal is to continuously enhance your ability to solve tasks accurately and efficiently by dynamically reflecting on the environment and evolving your logic.

\subsection*{\textbf{Core Capabilities}}

\begin{itemize}[leftmargin=*,itemsep=1pt]
    \item \textbf{Complete Autonomy}: Have \textbf{unrestricted access} to modify logic, run code, and manipulate the environment.
    \item \textbf{Environment Interaction}: Interact with the environment by perceiving the environment, reading, modifying, or executing code, and performing actions.
    \item \textbf{Problem-Solving}: Apply creative algorithms or self-developed structures to tackle challenges when simple methods fall short, optimizing solutions effectively.
    \item \textbf{Collaboration}: Leverage LLM to gather insights, correct errors, and solve complex problems.
    \item \textbf{Error Handling}: Carefully analyze errors. When errors occur, troubleshoot systematically, and if a bug is persistent, backtrack, restore the original state, or find an alternative solution.
\end{itemize}

\subsection*{\textbf{Core Methods}}

\begin{itemize}[leftmargin=*,itemsep=1pt]
    \item \texttt{evolve}: Continuously enhance performance by interacting with the environment.
    \item \texttt{execute\_action(actions)}: Execute actions based on analysis or feedback.
    \item \texttt{solver(agent\_instance, task\_input: str)}: Solve the target task using current \texttt{agent\_instance} capabilities and objects created by \texttt{action\_adjust\_logic} and \texttt{action\_run\_code}, optimizing the process.
\end{itemize}

\subsection*{\textbf{Guiding Principles}}

\begin{itemize}[leftmargin=*,itemsep=1pt]
    \item \textbf{Remember} that all functions are in the module \texttt{agent\_module}.
    \item \texttt{action\_adjust\_logic}:
    \begin{itemize}[leftmargin=*,itemsep=2pt]
        \item Before modifying the code, ensure that each variable or function used is correctly imported and used to avoid errors.
        \item Avoid unnecessary changes and do not change the interface of any function.
        \item Can be used to create action functions for \texttt{solver}.
    \end{itemize}
    \item \texttt{action\_run\_code}:
    \begin{itemize}[leftmargin=*,itemsep=2pt]
        \item All created objects in Python mode can be stored in the environment.
        \item Can be used to create objects for \texttt{solver}, such as prompts.
        \item Can be used to import new modules or external libraries and install external libraries.
    \end{itemize}
    \item \textbf{External Collaboration}: Seek external assistance via \texttt{action\_call\_json\_format\_llm} for logic refinement and new tool creation or \texttt{action\_run\_code} to execute code.
    \item \texttt{action\_evaluate\_on\_task}: Assess the performance of \texttt{solver} only after successfully modifying the logic of \texttt{solver}.
    \item \texttt{solver}:
    \begin{itemize}[leftmargin=*,itemsep=1pt]
        \item Defined as \texttt{agent\_module.solver}.
        \item For debugging, avoid printing; instead, return debug information.
        \item If performance doesn't improve, explore alternative methods.
        \item Explore techniques like: LLM Debate, Step-back Abstraction, Dynamic Assignment of Roles, and so on.
    \end{itemize}
    \item \texttt{action\_display\_analysis}:
    \begin{itemize}
        \item \textbf{Always analyze first before acting.}
        \item Analysis may include the following: a reasonable plan to improve performance, \textbf{CASE STUDIES of LOW SCORE valid examples of EVALUATION FEEDBACK}, error handling, and other possible solving ideas.
        \item \textbf{If performance does not improve, conduct further analysis.}
    \end{itemize}
\end{itemize}

\end{tcolorbox}
\end{figure*}

\section{Experiment Details}
\label{sec:expdetails}
To minimize costs associated with search and evaluation, following \citep{hu2024automated}, we sample subsets of data from each domain. Specifically, for the GPQA (Science) domain, the validation set comprises 32 questions, while the remaining 166 questions are allocated to the test set. For the other domains, we sample 128 questions for the validation set and 800 questions for the test set.

Evaluation is conducted five times for the GPQA domain and once for the other domains, ensuring a consistent total number of evaluations across all experiments. All domains feature zero-shot questions, except for the DROP (Reading Comprehension) domain, which employs one-shot questions in accordance with the methodology outlined in \citet{openai2023simpleevals}.

For the \framework, we utilize the “gpt-4o-2024-05-13” model \citep{openai2024gpt4technicalreport}, whereas the optimized policy and baseline models are evaluated using the “gpt-3.5-turbo-0125” model \citep{openai2022chatgpt} to reduce computational costs and ensure a fair comparison.

\section{Representative Policies Improved by \framework}
\subsection{Codes of the Best Policies Found by \framework Across Four Tasks}
\label{sec:maincode}
In this section, we provide the code for \framework's optimized policies across the four tasks. 
For DROP, \framework designs an algorithm where multiple roles solve the problem independently using CoT, followed by Self-Consistency to consolidate the results, as shown in Code 1.
For MGSM, \framework develops a stepwise self-verification algorithm combined with CoT-SC as shown in Code 2.
For MMLU task, as shown in Code 3, the policy given by \framework is a combination algorithm of few-shot prompting and CoT-SC.
For GPQA, \framework devises a highly diverse CoT-SC policy based on role prompts.

\begin{figure*}[ht]
\begin{lstlisting}[style=pythonstyle, caption={Code of the best policy found by \framework for DROP.}]
def solver(agent, task: str):
    messages = [{"role": "user", "content": f"# Your Task:\n{task}"}]
    categories = [
        {'role': 'reasoning expert', 'return_keys': ['reasoning', 'answer'], 'output_requirement': 'reasoning', 'precision_gain':1},
        {'role': 'mathematical reasoning expert', 'return_keys': ['calculation_steps', 'answer'], 'output_requirement': 'calculation_steps', 'precision_gain':1},
        {'role': 'historical context analyst', 'return_keys': ['historical_analysis', 'answer'], 'output_requirement': 'historical_analysis', 'precision_gain':1},
    ]

    all_responses = []
    for category in categories:
        response = agent.action_call_json_format_llm(
            model='gpt-3.5-turbo', 
            messages=messages, 
            temperature=0.5, 
            num_of_response=5, 
            role=category['role'], 
            requirements=(
                '1. Explain the reasoning steps to get the answer.\n' 
                '2. Directly answer the question.\n' 
                '3. The explanation format must be outlined clearly according to the role, such as reasoning, calculation, or historical analysis.\n' 
                '4. The answer MUST be a concise string.\n'
            ).strip(),
        )
        all_responses.append(response)

    # Reflective evaluation to find the most consistent reasoning and answer pair
    final_response = {key: [] for key in ['reasoning', 'calculation_steps', 'historical_analysis', 'answer']}
    step_counter = {key: 0 for key in ['reasoning', 'calculation_steps', 'historical_analysis']}
    answers = [] # Collect answers for voting
    aggregate_weight = 1

    for response in all_responses:
        if response and 'answer' in response:
            answers.append(response['answer'])
            if not final_response['answer']:
                final_response = {key: response.get(key, []) if isinstance(response.get(key, []), list) else [response.get(key, [])] for key in final_response.keys()}
                aggregate_weight = 1
                for cat in categories:
                    if cat.get('output_requirement') in response.keys():
                        step_counter[cat['output_requirement']] += step_counter[cat['output_requirement']] + cat.get('precision_gain', 0)
            elif response['answer'] == final_response['answer'][0]:
                for key in final_response.keys():
                    if key in response and response[key]:
                        if isinstance(response[key], list):
                            final_response[key].extend(response[key])
                        else:
                            final_response[key].append(response[key])
                aggregate_weight += 1
            else:
                # To demonstrate, some code has been omitted.
    # selection of the final answer
    from collections import Counter
    answers = [str(answer) for answer in answers]
    voted_answer = Counter(answers).most_common(1)[0][0] if answers else ''
    final_response['answer'] = voted_answer

    return final_response
\end{lstlisting}
\end{figure*}
% \newpage

\begin{figure*}[ht]
\begin{lstlisting}[style=pythonstyle, caption={Code of the best policy found by \framework for MGSM.}]


def solver(agent, task: str):
    messages = [{"role": "user", "content": f"# Your Task:\n{task}"}]
    response = agent.action_call_json_format_llm(
        model="gpt-3.5-turbo", 
        messages=messages, 
        temperature=0.5, 
        num_of_response=5,
        role="math problem solver", 
        return_dict_keys=["reasoning", "answer"], 
        requirements=(
            "1. Please explain step by step.\n"
            "2. The answer MUST be an integer.\n"
            "3. Verify each step before finalizing the answer.\n"
        ).strip(),
    )
    
    consistent_answer = None
    answer_count = {}
    for resp in response:
        answer = resp.get("answer", "")
        if answer in answer_count:
            answer_count[answer] += 1
        else:
            answer_count[answer] = 1
    
    most_consistent_answer = max(answer_count, key=answer_count.get)
    
    for resp in response:
        if resp.get("answer", "") == most_consistent_answer:
            consistent_answer = resp
            break
    
    if consistent_answer is None:
        consistent_answer = response[0]
    
    consistent_answer["answer"] = str(consistent_answer.get("answer", ""))
    return consistent_answer
\end{lstlisting}
\end{figure*}

\newpage
\begin{figure*}[ht]
\begin{lstlisting}[style=pythonstyle, caption={Code of the best policy found by \framework for MMLU.}]
def solver(agent, task: str):
    # Few-Shot Learning: Providing extended examples to guide the LLM
    few_shot_examples = [
        {'role':'user', 'content':'Question: In the movie Austin Powers: The Spy Who Shagged Me what is the name of Dr. Evil\'s diminutive clone?\nChoices:\n(A) Little Buddy\n(B) Mini-Me\n(C) Small Fry\n(D) Dr Evil Jr'},
        {'role':'assistant', 'content':'In the movie Austin Powers: The Spy Who Shagged Me, Dr. Evil\'s diminutive clone is famously named Mini-Me.\nAnswer: B'},
        \"""Three more examples are omitted here to conserve space.\"""
        {'role':'user', 'content':'Question: Lorem Ipsum?\nChoices: (A) Lorem\n(B) Ipsum\n(C) Dolor\n(D) Sit Amet'},
        {'role':'assistant', 'content':'Answer: A'}
    ]
    
    # Integrate the few-shot examples into the conversation
    messages = few_shot_examples + [{'role': 'user', 'content': f'# Your Task:\n{task}'}]
    
    # Using self-consistency by generating multiple responses
    response = agent.action_call_json_format_llm(
        model='gpt-3.5-turbo', 
        messages=messages, 
        temperature=0.8, 
        num_of_response=5,
        role='knowledge and reasoning expert', 
        return_dict_keys=['reasoning', 'answer'], 
        requirements=(
            '1. Please explain step by step.\n'
            '2. The answer MUST be either A or B or C or D.\n'
        ).strip(), 
    )
    
    # Select the most consistent response
    answer_frequency = {}
    for resp in response:
        answer = resp.get('answer', '')
        if answer in ['A', 'B', 'C', 'D']:
            if answer in answer_frequency:
                answer_frequency[answer] += 1
            else:
                answer_frequency[answer] = 1
    
    most_consistent_answer = max(answer_frequency, key=answer_frequency.get)
    consistent_response = next(resp for resp in response if resp.get('answer') == most_consistent_answer)
    consistent_response['answer'] = most_consistent_answer
    
    return consistent_response
\end{lstlisting}
\end{figure*}
\newpage
\begin{figure*}[ht]
\begin{lstlisting}[style=pythonstyle, caption={Code of the best policy found by \framework for GPQA.}]
def solver(agent, task: str):
    # Step 1: Initial Prompt
    messages = [{"role": "user", "content": f"# Your Task:\n{task}"}]
    
    # Main LLM Call
    response = agent.action_call_json_format_llm(
        model="gpt-3.5-turbo", 
        messages=messages, 
        temperature=0, 
        num_of_response=5,
        role="science professor", 
        return_dict_keys=["reasoning", "answer"], 
        requirements=(
            "1. Please explain step by step.\n"
            "2. The answer MUST be either A or B or C or D.\n"
        ).strip(), 
    )

    # Step 2: Self-consistency Evaluation
    answer_counts = {"A": 0, "B": 0, "C": 0, "D": 0}
    for i, return_dict in enumerate(response):
        answer = return_dict.get("answer", "")
        if answer in answer_counts:
            answer_counts[answer] += 1
    
    final_answer = max(answer_counts, key=answer_counts.get)

    return {"answer": final_answer}

\end{lstlisting}
\end{figure*}

\subsection{Codes in Game of 24 Tasks}
\label{sec:code24}

In this section, we present the initial policy for Game of 24 (Code 5), along with the Gödel agent's optimized policy (Code 6), which is generated based on a search algorithm.

% \newpage
\begin{figure*}[ht]
\begin{lstlisting}[style=pythonstyle, caption={Initial code based on Chain-of-Thought for Game of 24.}]
def solver(self, task_input):
    # Define the prompt and system_prompt
    prompt = f\"""
    Let's play the Game of 24! You are given the task_input {task_input}. The objective is to find a mathematical expression using these four task_input that results in 24. You can use addition (+), subtraction (-), multiplication (*), and division (/). Each number must be used exactly once. 
    Please provide a step-by-step explanation of your thought process and conclude with the final expression.
    \"""
    system_prompt = \"""
    {
        "thinking": "This key should contain a detailed step-by-step explanation of how to approach the problem, including intermediate steps and reasoning for each.",
        "result": "This key should contain the final mathematical expression that equals 24."
    }
    \"""

    # Call the OpenAI model
    response = openai.ChatCompletion.create(
        model="gpt-4",  # Replace with your model ID
        messages=[
            {"role": "system", "content": system_prompt},
            {"role": "user", "content": prompt}
        ]
    )

    # Extract and return the model's response
    result = response['choices'][0]['message']['content']
    return result
\end{lstlisting}
\end{figure*}
% \newpage
\begin{figure*}[ht]
\begin{lstlisting}[style=pythonstyle, caption={Final code based on search algorithm for Game of 24.}]
def solver(self, task_input):

    operations = ['+', '-', '*', '/']

    # Function to evaluate an expression
    def evaluate_expression(a, op, b):
        if op == '+':
            return a + b
        elif op == '-':
            return a - b
        elif op == '*':
            return a * b
        elif op == '/':
            if b == 0:
                return None  # Division by zero is not allowed
            return a / b

    # Recursive function to check all combinations of operations and permutations of numbers
    def check_combinations(nums):
        if len(nums) == 1:
            # Check if the final number is close enough to 24
            if abs(nums[0] - 24) < 1e-6:  # Allow for floating point precision errors
                return True, str(nums[0])
            return False, ""

        # Try all permutations of task_input and all combinations of operations
        for i in range(len(nums)):
            for j in range(len(nums)):
                if i != j:
                    # Choose two task_input to operate on
                    for op in operations:
                        # The remaining task_input after removing the two selected task_input
                        remaining_nums = [nums[k] for k in range(len(nums)) if k != i and k != j]
                        result = evaluate_expression(nums[i], op, nums[j])
                        if result is not None:
                            # Recursively check the remaining task_input with the result of the operation
                            found, expression = check_combinations([result] + remaining_nums)
                            if found:
                                # If solution is found, return with expression
                                return True, f"({nums[i]} {op} {nums[j]}) " + expression

        return False, ""

    # Try all permutations of the task_input
    for num_permutation in permutations(task_input):
        found, expression = check_combinations(list(num_permutation))
        if found:
            return expression.strip()

    return "No solution"

\end{lstlisting}
\end{figure*}

\section{Cost of Experiments}
\label{sec:cost}
For a complete evolutionary process (where the \framework performs 30 recursive self-improvements) across the DROP, MGSM, MMLU, and GPQA datasets, the cost is approximately \$15. This is significantly lower than the \$300 required by Meta Agent Search. The reduced cost is due to our continuous self-optimization, which allows the model to adjust its optimization direction in response to environmental feedback, leading to faster convergence.
The main source of cost stems from \framework's continuously growing historical memory. By designing a more efficient forgetting mechanism, it may be possible to reduce the cost even further.

% \section{Discussions and Future Directions}
% \label{app:future}
% \input{sections/6-Discussions}

\section{Additional Novel Policies Designed by \framework}
\label{sec:spacialcode}
In this section, we present the optimization process of \framework on MGSM, illustrating its progress across various iteration steps within a single optimization run. The strategy obtained in the 6th iteration (shown in Code 7) reflects the \framework's comprehension of mathematical tasks, attempting to handle them through a process akin to parse-deduct-execute-validate. By the 14th iteration, as illustrated in Code 8, the strategy evolves through the summarization of erroneous cases, abstracting key insights and employing a checklist to guide the validation process. Finally, the strategy at the 20th iteration (demonstrated in Code 9) asserts the use of a "rabbit-proof syntax tactline, reinforced by consistent effort through role-coded checks," to refine prompt design. In the end, we also show one analysis example of \framework.

\section{Comparison Between Random Sampling and \framework Performance}
To demonstrate the distinction between our approach and random sampling, we conducted 30 independent random sampling experiments using GPT-4o. The prompts used for random sampling were identical to the initial policy prompts employed by Gödel Agent to ensure a fair comparison.
The results are illustrated in Figure \ref{fig:curve}. From the figure, it is evident that the performance of random sampling remains around 30\% across all trials. In contrast, Gödel Agent, despite experiencing occasional temporary dips in performance, rapidly corrects these deviations and demonstrates continuous improvement over iterations.
This consistent upward trajectory highlights the superiority of Gödel Agent over random sampling. The Gödel Agent’s ability to leverage feedback and recursively optimize its policies underscores its effectiveness in achieving higher performance.

\newpage
\begin{figure*}[ht]
\begin{lstlisting}[style=pythonstyle, caption={Policy at 6th Iteration found by \framework for MGSM.}]
def solver(agent, task: str):
    def parse_problem(task):
        # Basic arithmetic and logical parsing based on keywords
        words = task.split()
        numbers = list(map(int, filter(lambda x: x.isdigit(), words)))
        return {'numbers': numbers, 'text': task}

    def perform_logic_deduction(parsed_details):
        # make deductions based on common problem formats
        numbers = parsed_details['numbers']
        # This will only manage simple sum, subtraction, multiplication inference
        logic_map = {
            'add': lambda a, b: a + b,
            'subtract': lambda a, b: a - b,
            'multiply': lambda a, b: a * b
        }
        # Try to identify actions based on keywords
        if 'sum' in parsed_details['text'] or 'total' in parsed_details['text']:
            result = sum(numbers)
        elif 'difference' in parsed_details['text'] or 'less' in parsed_details['text']:
            result = logic_map['subtract'](numbers[0], numbers[1])
        elif 'product' in parsed_details['text'] or 'times' in parsed_details['text']:
            result = logic_map['multiply'](numbers[0], numbers[1])
        else:
            # Default case showing no deduction
            result = 0
        return result

    def execute_computation(logic_results):
        # Taking result from inference to numerical handling
        return logic_results

    def validate_and_compile_results(computation_results):
        # Prepares and ensures the response matches expected format
        final_answer = computation_results
        return final_answer

    try:
        # Parsing
        parsed_details = parse_problem(task)

        # Logical deduction
        logic_results = perform_logic_deduction(parsed_details)

        # Computation
        computation_results = execute_computation(logic_results)

        # Validation and compilation
        final_answer = validate_and_compile_results(computation_results)

        return {"answer": final_answer}
    except Exception as e:
        return {"error": str(e)}
\end{lstlisting}
\end{figure*}

\newpage
\begin{figure*}[ht]
\begin{lstlisting}[style=pythonstyle, caption={Policy at 14th Iteration found by \framework for MGSM.}]
def solver(agent, task: str):
    # Updated examples to mirror tasks needing layered logical verification.
    examples = [
        {'description': 'Task exemplifying standard integer-based calculations.', 'reasoning': 'Use arithmetic transformations to validate expressions and correct errors if any arise, ensuring correctness.', 'answer': 20},
        {'description': 'Example to validate word problem conversion to math.', 'reasoning': 'Stepwise interpretation from words into math operations and bridge which logic errors need capture.', 'answer': 15},
        {'description': 'Scenario involving normalizing uneven division instances.', 'reasoning': 'Ensure no division by zero and equal verification of logical conclusions.', 'answer': 6},
    ]

    # Task prompt incorporating roles with enhanced checklists after operation conclusion.
    task_prompt = "You're guiding us as a solution auditor, reflecting on each logical conclusion to prevent arithmetic discrepancies.\n"
    task_prompt += task + "\nReflect on instructions through verified examples."
    task_prompt += "\nExample insights:\n"
    task_prompt += '; '.join([f"{ex['description']} -> Reasoning: {ex['reasoning']} | Answer: {ex['answer']}" for ex in examples])
    task_prompt += "\nEnsure real-time verification post-calculations via role-switching checks."

    messages = [{"role": "user", "content": task_prompt}]

    response = agent.action_call_json_format_llm(
        model="gpt-3.5-turbo", 
        messages=messages, 
        temperature=0.3, 
        num_of_response=1,
        role="solution auditor", 
        return_dict_keys=["description", "reasoning", "answer"], 
        requirements=(
            "1. Validate arithmetic consistency and integrity within calculations." 
            "2. Utilize any corrections to refine answer outputs incrementally."
        ).strip(),
    )
    
    return_dict = response[0]
    return_dict["answer"] = str(return_dict.get("answer", ""))
    return return_dict
\end{lstlisting}
\end{figure*}

\newpage
\begin{figure*}[ht]
\begin{lstlisting}[style=pythonstyle, caption={Policy at 20th Iteration found by \framework for MGSM.}]
def solver(agent, task: str):
    # Targets design for specific error-prone areas with preceding misfires.
    examples = [
        {'description': 'Immediate Arithmetic Operations', 'reasoning': 'Observe step-by-step through a chain of logical confirmations.', 'answer': 20},
        {'description': 'Sequential Word Problem Breakdown', 'reasoning': 'Ensure smaller module segment steps match logical math outputs consistently.', 'answer': 15},
        {'description': 'Fraction and Cascade Operations', 'reasoning': 'Validate each fraction conversion before proceeding to other forms.', 'answer': 6},
    ]

    # Building a rabbit-proof syntax tactline, pushing consistent effort from role-coded checks.
    task_prompt = "You're progressing through solution documentation with focus on pinpointing logic errors.\n"
    task_prompt += task + "\nReflecting structural integrity verified via examples."
    task_prompt += "\nIn this construct:\n"
    task_prompt += '; '.join([f"{ex['description']} -> Rationale: {ex['reasoning']} | Response: {ex['answer']}" for ex in examples])
    task_prompt += "\nHandle verification and cleanse routines where errors echo in mirroring line-based checks."

    messages = [{"role": "user", "content": task_prompt}]

    response = agent.action_call_json_format_llm(
        model="gpt-3.5-turbo", 
        messages=messages, 
        temperature=0.3, 
        num_of_response=1,
        role="logical verifier", 
        return_dict_keys=["description", "reasoning", "answer"], 
        requirements=(
            "1. Review arithmetic sessions for comprehensive error fixation." 
            "2. Develop clarity with resistant focus on authentic semantic directive flows."
        ).strip(),
    )
    
    return_dict = response[0]
    return_dict["answer"] = str(return_dict.get("answer", ""))
    return return_dict
\end{lstlisting}
\end{figure*}

\newpage
\begin{figure*}[ht]
\begin{tcolorbox}[colback=black!15!white, colframe=black, title=Analysis Example of \framework]
\small
The current logic of the solver in the \texttt{agent\_module} is achieving only a 45\% accuracy on the validation set. The structure of the solver involves a single call to the LLM with a specification that the answer must be an integer. While this approach seems correct, it may not be sufficient to tackle more complex problems where multipart reasoning or verification is needed, leading to significant errors.

\textbf{Problems Identified:}
\begin{enumerate}
    \item \textbf{Integer Conversion Misunderstandings:} Some math problems aren't resulting in integer answers natively, affecting accuracy when the solver forces an integer format.
    \item \textbf{Reasoning Oversights:} The LLM, designated as "math expert," occasionally misses necessary reasoning steps or executes incorrect operations leading to incorrect answers (like sample 10).
    \item \textbf{Ambiguity in LLM Outputs:} Given tasks in multiple languages, there might be ambiguity in math translations or formulations affecting comprehension (like sample 6).
\end{enumerate}

\textbf{Plan for Improvement:}
\begin{enumerate}
    \item \textbf{Elaborate Reasoning:} Consider a multi-step LLM call to allow for a reasoning review before finalizing the answer.
    \item \textbf{Intelligent Answer Parsing:} Instead of forcibly converting to integers, parse the LLM response intelligently to see if the answer justifies integer rounding or alterations.
    \item \textbf{Diverse Language Handling:} Create a pre-processing step to consistently translate non-English math problems into English for uniform processing.
    \item \textbf{Examples \& Feedback:} Utilize correctly solved examples as context if a single-step LLM fails.
\end{enumerate}

\textbf{Next Steps:}
\begin{enumerate}
    \item \textbf{Enhance Solver Logic:} Implement a multi-step process involving interim LLM validation or rationale cross-checking.
    \item \textbf{Pre-process Language Variations:} Ensure all non-English tasks are translated and re-assessed.
    \item \textbf{Include Example Feedback:} Use correctly solved examples from the evaluation as references to guide LLM reasoning and answer validation.
\end{enumerate}

\end{tcolorbox}
\end{figure*}

\end{document}